\documentclass[letterpaper, 10 pt, conference]{ieeeconf}  

\IEEEoverridecommandlockouts                              

\usepackage{amsfonts}    
\usepackage{bm}
\usepackage {mathtools}
\usepackage{graphicx} 
\usepackage{subfigure}
\usepackage{amsmath, amssymb}
\usepackage{makecell}
\usepackage{multirow}
\usepackage{makecell}
\usepackage{amsmath}
\usepackage{arydshln}
\usepackage{multirow}

\usepackage{booktabs}
\usepackage{mathrsfs}
\usepackage{array}

\title{\LARGE \bf
Point-line-based RGB-D SLAM and Bundle Adjustment Uncertainty Analysis 
}

\author{Xin Ma and Xinwu Liang 
\thanks{The authors are with School of Aeronautics and Astronautics,
        Shanghai Jiao Tong University, Shanghai 200240, China
        {\tt\small maxin1900@sjtu.edu.cn;xinwuliang@sjtu.edu.cn}}%
}

\begin{document}

\maketitle
\thispagestyle{empty}
\pagestyle{empty}

\begin{abstract}

 Most of the state-of-the-art indirect visual SLAM methods are based on the sparse point features. However, it is hard to find enough reliable point features for state estimation in the case of  low-textured scenes. Line features are abundant in urban and indoor scenes. Recent studies have shown that the combination of point and line features can provide better accuracy despite the decrease in computational efficiency. In this paper, measurements of point and line features are extracted from RGB-D data to  create map features, and points on a line are treated as keypoints. We propose an extended approach to make more use of line observation information. And we prove that, in the local bundle adjustment, the estimation uncertainty of keyframe poses can be reduced when considering more landmarks with independent measurements in the optimization process. Experimental results on two public RGB-D datasets demonstrate that the proposed method has better robustness and accuracy in challenging environments.

\end{abstract}


\section{INTRODUCTION}

Visual odometry (VO)  and Simultaneous Localization and Mapping (SLAM) are popular topics in the field of robotics research. In recent years, they receive more attention due to their applications in self-driving cars, augmented reality and 3D reconstruction. Visual odometry and SLAM can be addressed with different camera sensors, such as monocular cameras \cite{forster2014svo}, stereo cameras \cite{ pire2015stereo}, and RGB-D cameras \cite{kerl2013robust}. RGB-D cameras can provide depth measurements for each frame and reduce costs for indoor state estimation and mapping despite of their relatively short depth measurement range. Generally, lack of features and uneven feature distributions can pose challenges for feature-based visual SLAM methods. Therefore, in this work, we aim at developing a more robust and accurate RGB-D SLAM approach.

Current visual SLAM methods can be typically divided into two categories, indirect (or feature-based) such as PTAM \cite{klein2007parallel} and ORB-SLAM \cite{mur2015orb} and direct methods such as \cite{engel2014lsd,engel2017direct}. Indirect methods usually need to extract features from the image and then match them with history features for data association, so that a 3D sparse map can be built and the pose of subsequent frames can be calculated by minimizing geometric errors of the features in the image plane. Direct methods directly exploit on the pixel intensity and estimate the pose by minimizing photometric errors of a certain amount of pixels, while avoiding the processes of feature extraction and matching. 

Among the indirect methods, most visual SLAM systems are based on point features because they are easy to extract with less time consumption. However, these methods rely on sufficient keypoint distributions in frames and are vulnerable to the low-textured scenes. The lack of reliable feature points often occurs in man-made environments where a certain number of lines may still exist. Methods with the combination of point and line features have been proposed in recent years \cite{gomez2019pl,pumarola2017pl,yang2017direct}. Point-line-based methods showed better performance than point-based methods in some low-textured scenes. In addition, most of current point-line-based methods represent a 3D line by its 3D endpoints \cite{pumarola2017pl} or the $Pl\ddot{u}cker$ coordinates with six degrees of freedom \cite{zuo2017robust}. 

In order to use more measurement information of line features from the RGB-D data, we sample a set of points on the line segments to create map features, and then compute the reprojection errors based on the distance between points and lines. The line segments are used to guide the selection of points and construct reprojection errors for these points. In summary, the main contributions of this paper are: 
\begin{itemize}
	\item Under the maximum likelihood estimation and Gaussian noise assumption, we prove that including more landmarks with independent measurements in the local bundle adjustment can produce less uncertainty in the estimation of local keyframe poses. The significance of our theoretical analysis is that, we provide a heuristic example to explain  why the more observation information used in a SLAM system, the more accurate the pose estimation usually. The analysis is not limited to point features and RGB-D cameras. 
	
	\item A robust RGB-D SLAM method is presented by combining point and line guidance features. A map line is represented by multiple points on the line instead of endpoints or the $Pl\ddot{u}cker$ coordinates. Experiments are performed on public RGB-D datasets to evaluate the good performance of our method, compared with other existing methods.
\end{itemize}

\section{RELATED WORK}

Our algorithm focuses on how to make better use of line features in SLAM systems, which belongs to indirect methods. Thus, in this section, we divide the recent related VO or SLAM works into two categories, point-based and point-line-based methods, and briefly introduce them.

\subsection{Point-based Methods}

Many famous visual odometry and SLAM works have been developed based on point features, for example PTAM and ORB-SLAM. In these works, keypoints are extracted and exploited for pose estimation and mapping with different descriptors. Then the keypoints in different frames are matched according to their descriptor distances. Based on the feature correspondences, a sparse 3D map is built so that the subsequent camera poses are estimated by solving the PnP  problems. Methods based on monocular cameras inevitably encounter the problem of scale drift due to depth ambiguity. In contrast, RGB-D and stereo cameras can easily avoid this issue \cite{steinbrucker2011real}, but stereo camera methods cost more time due to the feature matching between left and right images. In terms of robustness,  low-textured scenes are considered as a main challenge for indirect methods.

Compared with feature-based methods, direct approaches \cite{engel2013semi} directly deal with the raw pixel intensities, which can significantly save the time of feature extraction and matching. 
Direct methods have been performed for different sensors, such as LSD-SLAM for monocular cameras and DVO \cite{kerl2013dense} for RGB-D cameras. High computational efficiency is one of the advantages of direct methods. For example, as a typical semi-dense method, SVO \cite{forster2014svo} can run with hundreds of frames per second. The core of these methods is the minimization of the photometric errors for pixels of a certain size, so they can operate well in some texture-less scenes with few point features. 

\subsection{Point-line-based Methods}

Considering that the detection of line features is less sensitive to lighting variations and the accuracy of line-based methods is usually not comparable with that of point features, combinations of point and line features have been proposed for VO/SLAM methods recently. For feature-based methods, lines are usually treated similarly to points, that is, the line features are detected first and then the corresponding descriptors are computed for feature matching \cite{pumarola2017pl}. The work \cite{gomez2019pl} proposed a stereo point-line-based SLAM method with open source code, and lines are integrated into the process of loop closing. More recently, PL-VIO is presented in the work \cite{he2018pl}, which combines point and line features in a visual inertial system. 

Besides the indirect methods, line features can also be integrated into direct methods. Based on the idea that the points on a line have large pixel gradient, DLGO \cite{li2018direct} used a set of points in a line rather than two endpoints, which improved the performance of DSO.

In addition, the work \cite{sola2012impact} discussed the combined use of point and line features with several parameterizations for an EKF-SLAM system. Point-line-based methods have also been proposed for RGB-D cameras.

Lu et al. \cite{lu2015robust} proposed a RGB-D visual odometry and analyzed the uncertainty of motion estimation to show the benefit of fusing line features with point features. 
In terms of the relationship between features and the accuracy of pose estimation, the work in \cite{strasdat2010real} has conducted experiments and shown that increasing the number of features leads to higher accuracy of a monocular SLAM system. 

\section{SYSTEM OVERVIEW}

We build our system upon the ORB-SLAM2 \cite{mur2017orb} framework, which is a relatively complete SLAM system including relocalization, loop closing and map reuse capabilities (see Fig. 1). The proposed SLAM approach is also based on three main threads: tracking, local mapping and loop closing. For the details of system and point feature operations, ORB-SLAM and ORB-SLAM2 are suggested for reference. 

\begin{figure}[thpb]
	\centering
	\includegraphics[width=0.415\textwidth]{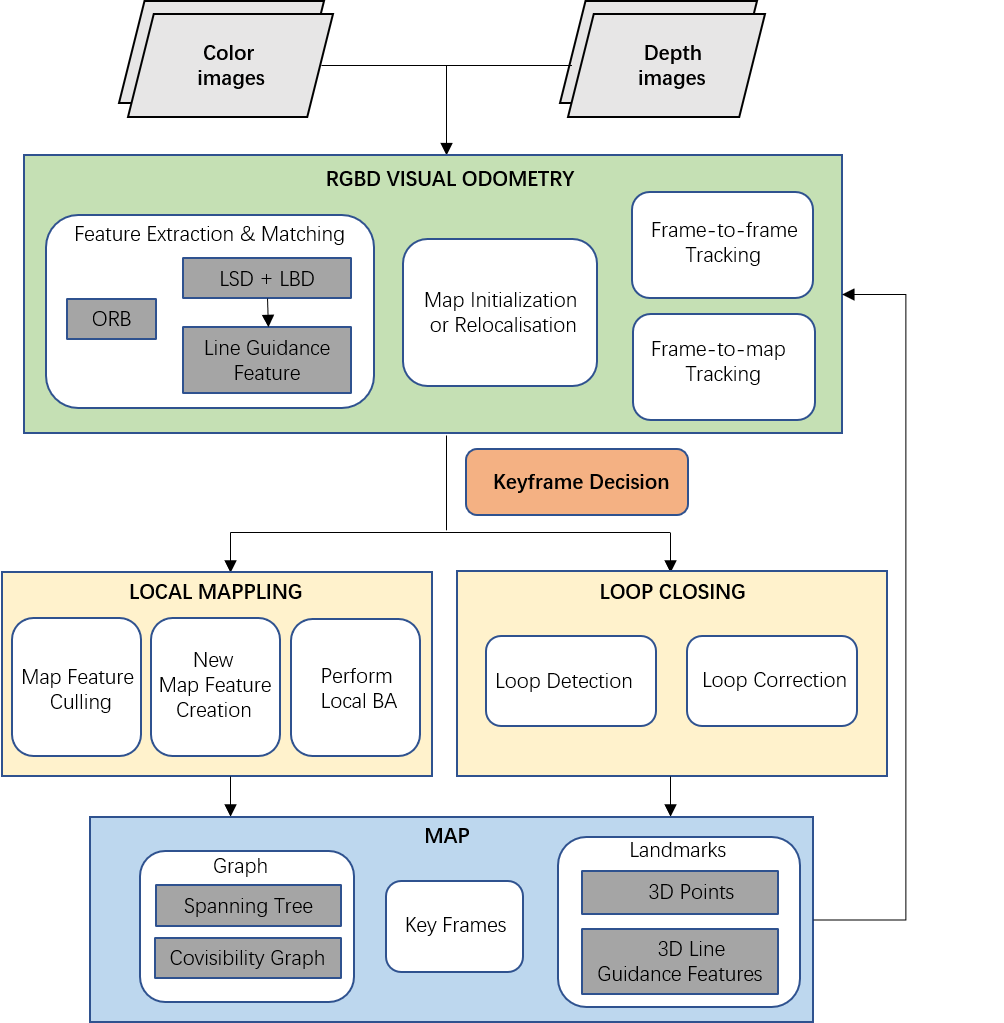}
	\caption{System overview, an extension of the ORB-SLAM2 pipeline.}
\end{figure}

The ORB detector is employed for extracting point features, while line segments are detected with LSD \cite{von2008lsd}, an O(n) line segment detector with good speed and quality. We compute the LBD descriptors \cite{zhang2013efficient} for each line and match them based on the distance between different descriptors. The LBD descriptor is considered to be robust against image artifacts without costing too much computational time. 

For RGB-D cameras, a 3D map is constructed and updated by back-projecting the tracked point and line features. The motion estimation is solved through a probabilistic Gauss-Newton minimization of the point and line geometric reprojection errors. In the process of optimization, we adopt a Pseudo-Huber loss function to remove the outliers and reduce the negative influence of feature mismatches. In addition, we  use line features for tracking and local mapping threads, preserving the original relocalization and loop closing modules of ORB-SLAM2. We review some issues related to the line features in section \uppercase\expandafter{\romannumeral4} and analyze the uncertainty of state estimation in the local bundle adjustment in section \uppercase\expandafter{\romannumeral5}.





\section{LINE-BASED ERROR}

In general, two 3D endpoints are used to parameterize a spatial line \cite{gomez2019pl,pumarola2017pl}, whose corresponding matching lines (i.e. measurements) can be found in the 2D image plane. Due to the characteristics of line matching, the endpoints of pairwise lines are not often strictly aligned. But the projection of the 3D line should be collinear with the matching line under the correct estimation.


For a 3D line, let $ \bm{P} $, $ \bm{Q} \in  \mathbb{R}^3$ be the homogeneous coordinates of 2D endpoints of its projection line in the image plane, $ \bm{P'} $, $ \bm{Q'} \in \mathbb{R}^2 $ be the endpoints of the 2D matching line. According to $ \bm{P'} $ and $ \bm{Q'} $ we can determine the corresponding 2D line equation, which has a line coefficient vector $ \bm{l}_{0} = {[a, b, c]}^\top $ and a normalized line coefficient vector: 
\begin{equation}
\bm{l} = \frac{ \bm{l}_0 }{ \sqrt{ a^2 + b^2}}.
\end{equation}

The line reprojection error $ E_{line} $ is defined based on the distance between the  projected endpoints and the 2D matching line:
\begin{equation}
E_{line} = E^2_{pl}(\bm{P}, \bm{l}) +  E^2_{pl}(\bm{Q}, \bm{l})
\end{equation}
with
\begin{equation}
 E_{pl}(\bm{P}, \bm{l}) = \bm{l}^\top \bm{P}.
\end{equation}

As shown in Fig. 2, we consider three different relationships between the projection of a 3D line  and the 2D matching line.      
Suppose that the midpoint ($P_{m}$) and the endpoints ($P_{s}$ and $P_{e})$ are not outliers, the larger the point-line distance error, the greater influence in the optimization problem. For example in the case of Fig. 2(b), the error $e_{3}$ formed by the point $P_{m}$ is larger than $e_{2}$, and it is more reasonable to use the midpoint $P_{m}$ to construct the line reprojection error instead of the endpoint $P_{e}$. Keeping this in mind, perhaps using more points on the same line to form the reprojection errors can provide more geometric constraints for the optimization problem and improve the system performance. In the next section we will analyze the uncertainty of pose estimation and show the benefit of combining line features and point features in a relatively complete SLAM system, which contains a local map optimization (i.e. local bundle adjustment).

\begin{figure}[htbp]
	\centering
	\subfigure[]{
		\begin{minipage}[t]{0.28\linewidth}
			\centering
			\includegraphics[width=1in]{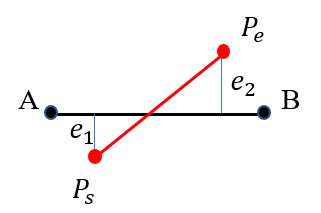}
		\end{minipage}
	}
	\subfigure[]{
		\begin{minipage}[t]{0.28\linewidth}
			\centering
			\includegraphics[width=1in]{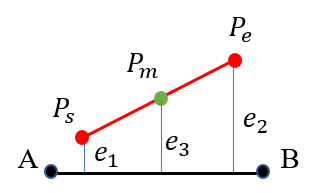}
		\end{minipage}
	}
	\subfigure[]{
	\begin{minipage}[t]{0.28\linewidth}
		\centering
		\includegraphics[width=1in]{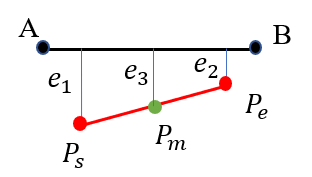}
	\end{minipage}
    }
	\centering
	\caption{Different situations when projecting 3D line to 2D plane. The red lines are the projection of 3D lines, whose corresponding 2D line are black lines. The endpoints of projected line are on either side of the matching line in $ \mathbf{(a)} $. The endpoints of projected line are on the same side of the matching line in $ \mathbf{(b)} $ and $ \mathbf{(c)} $ with different distances to the matching line. }
\end{figure}

\section{STATE ESTIMATION \& UNCERTAINTY ANALYSIS}

The work in \cite{lu2015robust} has proved that the combination of point and line features can reduce the uncertainty of pose estimation between two frames than just using points or lines in a RGB-D visual odometry. In this paper, we extend the uncertainty analysis to the case of local bundle adjustment (BA), which jointly optimizes multiple keyframe poses and many landmark positions and is also the core of sliding window algorithms in some other systems. This analysis can explain the relationship between data association and estimation uncertainty in a way for a relatively general SLAM system.

In the local map, the variables to be optimized include the current keyframe $K_{i}$, the keyframes connected to $K_{i}$ in the covisibility graph and all the landmarks observed by these keyframes. The keyframes that can observe those landmarks but are not connected to $K_{i}$ (i.e. those share insufficient landmarks with $K_{i}$ ) remain fixed in the optimization.

%


The transformation matrix $ \bm{T(\bm{\xi})}  \in SE(3)$ can be represented by a six-vector $\bm{\xi} $ in the Lie algebra or by the rotation matrix $\bm{R}$ and translation vector $\bm{t}$:

\begin{equation}
\bm{T(X) = RX + t} 
\end{equation}
where $\bm{X} \in \mathbb{R}^3 $  is a 3D point in the world coordinates.

The projection function $\pi$ from 3D to 2D is defined as:

\begin{equation}
\pi\left(\begin{bmatrix}    
X_c \\
Y_c \\ 
Z_c
\end{bmatrix} \right) = 
\begin{bmatrix}
f_{x}\frac{X_{c}}{Z_{c}} + c_{x}  \\
f_{y}\frac{Y_{c}}{Z_{c}} + c_{y} 
\end{bmatrix} 
\end{equation}
where $(c_{x}, c_{y})$ is the principal point of camera, $ (f_{x}, f_{y}) $ is the focal length, all obtained from camera calibration.

\subsection{ Local BA with Original Landmarks }

We define the variables to be optimized in the local bundle adjustment as follows:
\begin{equation}
\bm{x} = {[ \bm{\xi}_{1}^\top, \cdots,\bm{\xi}_{m}^\top, \bm{P}_{1}^\top, \cdots, \bm{P}_{n}^\top ]}^\top  
\end{equation}
which contains $m$ keyframe poses and $n$ landmark positions in the local map. Let the number of fixed keyframes in the local map be $d$, and the current keyframe be $C_{0}$. 
If the $j$-th landmark is observed in the $i$-th keyframe, the corresponding reprojection error  $ \bm{e}_{ij}(0 \le i \le (m+d),1 \le j \le n) $ can be computed from $ \bm{\xi}_{i} $ and $\bm{P}_{j}$:
\begin{equation}
\bm{e}_{ij} = \pi \left( \bm{T}_{i}(\bm{P}_{j})  \right) - \bm{p}_{ij} 
\end{equation}
where $\bm{p}_{ij}$ is the observation of the landmark $\bm{P}_{j}$ on the $i$-th keyframe. Note that the reprojection error expression is different for different types of landmarks (e.g.,line landmarks). Next, we integrate all the reprojection errors in the local BA into an error function:

\begin{equation}
\label{h}
h( \bm{x} ) 
=
\begin{bmatrix}
\bm{e}_{01}  \\
\vdots     \\
\bm{e}_{ij}
\end{bmatrix}  
\end{equation}
where the form of $h( \bm{x} ) $ varies with the observation relationships in the local map. In order to better illustrate the problem, we have shown a factor graph corresponding to the local BA in Fig. 3 as an example. The corresponding error function in Fig. 3 is $h_{1}(\bm{x}) = {[ \bm{e}^\top_{01}, \bm{e}^\top_{02}, \bm{e}_{03}^\top, \bm{e}_{12}^\top, \bm{e}_{13}^\top, \bm{e}_{14}^\top, \bm{e}_{24}^\top, \bm{e}_{34}^\top]}^\top  $. For a more general representation, we define the error function as (\ref{h}). According to the Maximum Likelihood Estimation (MLE), the core of local bundle adjustment is to solve the following least square problem:
\begin{equation}
\mathop{\arg\min}_{\bm{x}}  h( \bm{x})^\top  \bm{\Sigma}^{-1}_{h}  h( \bm{x})
\end{equation}
where $\bm{\Sigma}_{h} = diag(\bm{\Sigma}_{01}, \cdots, \bm{\Sigma}_{ij})$ and $\bm{\Sigma}_{ij}$ is the covariance of the observation $\bm{p}_{ij}$. Note that (9) is derived based on the Maximum a Posteriori (MAP) problem, which is equivalent to the Maximum Likelihood Estimation (MLE) in a typical visual SLAM problem. And observations of the landmarks need to satisfy independence assumptions.

\begin{figure}[thpb]
	\centering
	\includegraphics[width=0.39\textwidth]{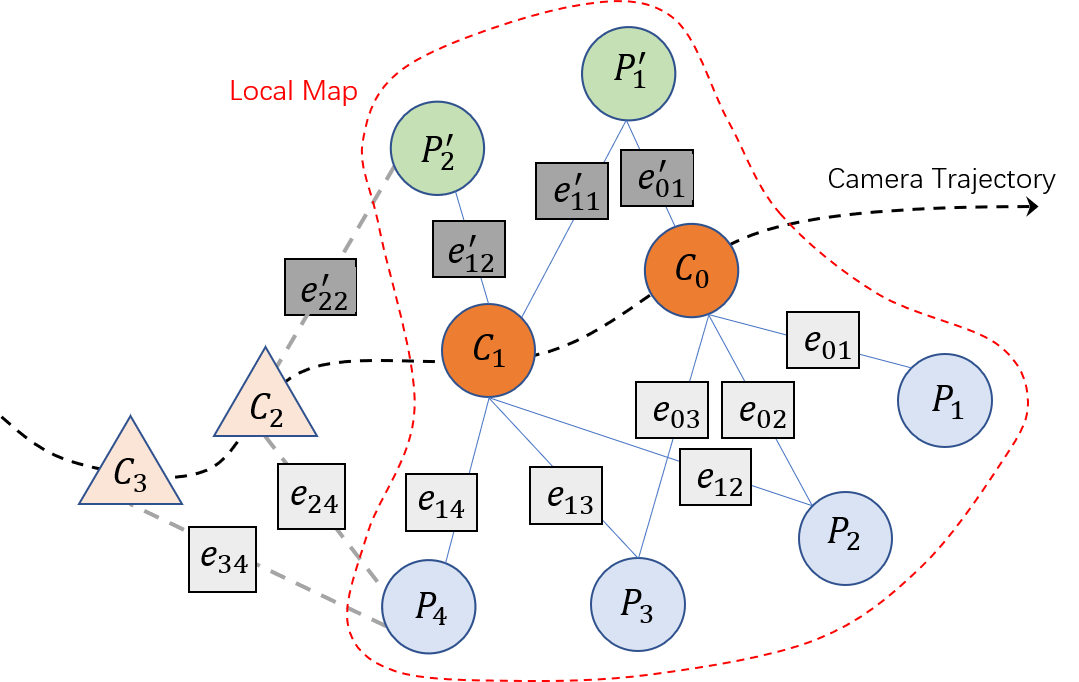}
	\caption{Factor graph for local bundle adjustment. The circles represent optimization variables, the light grey boxes represent error edge constraints for original landmarks, the dark grey boxes represent error edge constraints for new landmarks, and the triangles represent fixed keyframes. Note that $C_{0}$ is the current keyframe, $C_{2}$ and $C_{3}$ are the fixed keyframes, $C_{0}$ and $C_{1}$ are the keyframes to be estimated in the local BA. }
\end{figure}

Next, we define the following notations:
\begin{equation}
\bm{x}_{c} = {[ \bm{\xi}_{1}^\top, \cdots,\bm{\xi}_{m}^\top ]}^\top   \in \mathbb{R}^{6m},
 \\
\bm{x}_{p} = {[ \bm{P}_{1}^\top, \cdots,\bm{P}_{m}^\top ]}^\top   \in \mathbb{R}^{3n}
\end{equation}

According to first order linear approximation and the back-propagation of covariance, the MLE of $\bm{x}$ has covariance \cite{hartley2003multiple}:

\begin{equation}
cov(\bm{x})  =  \left( \bm{J}^\top_{h}  \bm{\Sigma}^{-1}_{h} \bm{J}_{h} \right)^{-1}   
\end{equation}
with $\bm{J}_{h} = \frac{\partial h }{\partial \bm{x} }  =  
\begin{bmatrix}
 \frac{\partial h }{\partial \bm{x}_{c} } & \frac{\partial h }{\partial \bm{x}_{p} }
\end{bmatrix}
=
\begin{bmatrix}
\bm{J}_{A} & \bm{J}_{B} 
\end{bmatrix}  $. We further derive $
\bm{J}^\top_{h}  \bm{\Sigma}^{-1}_{h} \bm{J}_{h} =
\begin{bmatrix}
\begin{array}{cc}
\bm{J}^\top_{A}\bm{\Sigma}^{-1}_{h}\bm{J}_{A} & \bm{J}^\top_{A}\bm{\Sigma}^{-1}_{h}\bm{J}_{B}  \\
\bm{J}^\top_{B}\bm{\Sigma}^{-1}_{h}\bm{J}_{A} & \bm{J}^\top_{B}\bm{\Sigma}^{-1}_{h}\bm{J}_{B}
\end{array}
\end{bmatrix} 
=
\begin{bmatrix}
	\begin{array}{cc}
		\bm{H}_{A} & \bm{H}_{B}  \\
		\bm{H}_{B}^\top & \bm{H}_{C} 
	\end{array}
\end{bmatrix}   $. 

Generally, the accuracy of camera trajectory gets more attention than geometrical reconstruction (e.g. landmarks), especially for the indirect SLAM methods.   As the first part of the state $\bm{x}$, the covariance of $\bm{x}_{c}$ corresponds to the upper left submatrix of $cov(\bm{x})$, with the size of $6m \times 6m$. Therefore, under Gaussian noise assumption, the MLE of  $\bm{x}_{c}$ (multiple keyframe poses in the local map) has covariance:
\begin{equation}
\label{C1}
\bm{C}_{h} = \left( \bm{H}_{A} - \bm{H}_{B} \bm{H}_{C}^{-1}  \bm{H}_{B}^\top  \right) ^{-1}
\end{equation}

\subsection{ Local BA with New Landmarks}

If we solve the problem of local BA with  $k$ new landmarks $\{\bm{P}'_{1}, \cdots, \bm{P}'_{k}\}$ and the same keyframe states as mentioned in Section \uppercase\expandafter{\romannumeral5}-A, we can define these new landmarks as:
\begin{equation}
\bm{x}_{p'} = {[ {\bm{P'}_{1}}^\top, \cdots, {\bm{P'}_{k}}^\top ]}^\top   \in \mathbb{R}^{3k}
\end{equation}

In this case, the variables to be optimized are written as:
\begin{equation}
\bm{x'} = {[ \bm{\xi}_{1}^\top, \cdots,\bm{\xi}_{m}^\top, {\bm{P'}_{1}}^\top, \cdots, {\bm{P'}_{k}}^\top ]}^\top 
= { [\bm{x}_{c}^\top, \bm{x}_{p'}^\top] }^\top 
\end{equation}
Let the number of fixed keyframes be $d'$ in this case. Define the error function as:
\begin{equation}
f( \bm{x'} ) 
=
\begin{bmatrix}
\bm{e'}_{01}  \\
\vdots     \\
\bm{e'}_{ij}
\end{bmatrix}  
\end{equation}
where $\bm{e'}_{ij} (0 \le i \le (m+d'),1 \le j \le k)$ represents the reprojection error corresponding to the $j$-th new landmark $\bm{P'}_{j}$ in the $i$-th keyframe:
\begin{equation}
\bm{e'}_{ij} = \pi \left(  \bm{T}_{i}(\bm{P'}_{j}) \right) - \bm{p}'_{ij} 
\end{equation}
For example, the error function can be expressed as $f_{1}(\bm{x}) = {[ \bm{e'}^{\top}_{01}, \bm{e'}^\top_{11}, \, \bm{e'}^\top_{12}, \bm{e'}^\top_{22}]}^\top $ in Fig. 3. 

The local BA solves the following least square problem:
\begin{equation}
\mathop{\arg\min}_{\bm{x'}}  f( \bm{x'})^\top  \bm{\Sigma}^{-1}_{f}  f( \bm{x'})
\end{equation}
where $\bm{\Sigma}_{f} = diag( \bm{\Sigma}'_{01}, \cdots, \bm{\Sigma}'_{ij})$ and $\bm{\Sigma}'_{ij}$ is the covariance of the observation $\bm{p}'_{ij}$.

According to the back-propagation of covariance, the MLE of $\bm{x'}$ has covariance:
\begin{equation}
cov(\bm{x'})  =  \left( \bm{J}^\top_{f}  \bm{\Sigma}^{-1}_{f} \bm{J}_{f} \right)^{-1}   
\end{equation}
where $\bm{J}_{f} = \frac{\partial f }{\partial \bm{x'} }  =  
\begin{bmatrix}
\frac{\partial f }{\partial \bm{x}_{c} } & \frac{\partial f }{\partial \bm{x}_{p'} }
\end{bmatrix}
=
\begin{bmatrix}
\bm{J}_{C} & \bm{J}_{D} 
\end{bmatrix}  $. We further derive $
\bm{J}^\top_{f}  \bm{\Sigma}^{-1}_{f} \bm{J}_{f} =
\begin{bmatrix}
\begin{array}{cc}
\bm{J}^\top_{C}\bm{\Sigma}^{-1}_{f}\bm{J}_{C} & \bm{J}^\top_{C}\bm{\Sigma}^{-1}_{f}\bm{J}_{D}  \\
\bm{J}^\top_{D}\bm{\Sigma}^{-1}_{f}\bm{J}_{C} & \bm{J}^\top_{D}\bm{\Sigma}^{-1}_{f}\bm{J}_{D}
\end{array}
\end{bmatrix} 
=
\begin{bmatrix}
\begin{array}{cc}
\bm{H}'_{A} & \bm{H}'_{B}  \\
\bm{H}'^\top_{B} & \bm{H}'_{C} 
\end{array}
\end{bmatrix}   $. 

Therefore, in this case, under Gaussian noise assumption, the MLE of $\bm{x}_{c}$ has covariance:
\begin{equation}
\label{C2}
\bm{C}_{f} = \left( \bm{H}'_{A} - \bm{H}'_{B} \bm{H}'^{-1}_{C}  \bm{H}'^\top_{B}  \right) ^{-1}
\end{equation}

\subsection{ Local BA with All the Landmarks}

If considering all the landmarks as mentioned in Section V-A and Section V-B, the states to be estimated in the local map is denoted as: 
\begin{equation}
\bm{x''} = {[ \bm{\xi}_{1}^\top, \cdots,\bm{\xi}_{m}^\top, \bm{P}_{1}^\top, \cdots, \bm{P}_{n}^\top, {\bm{P'}_{1}}^\top, \cdots, {\bm{P'}_{k}}^\top ]}^\top  
\end{equation}
The error function can be expressed as:
\begin{equation}
g( \bm{x''} ) 
=
\begin{bmatrix}
h(\bm{x})  \\ 
f(\bm{x'})
\end{bmatrix}  
\end{equation}
Local BA solves the following problem:
\begin{equation}
\mathop{\arg\min}_{\bm{x''}}  g( \bm{x''})^\top  \bm{\Sigma}^{-1}_{g}  g( \bm{x''})
\end{equation}
where $\bm{\Sigma}_{g} = diag(\bm{\Sigma}_{01}, \cdots, \bm{\Sigma}_{ij},  \bm{\Sigma}'_{01},  \cdots, \bm{\Sigma}'_{ij})$.
 
The MLE of $\bm{x''}$ has covariance: \\
\begin{equation}
cov(\bm{x''})  =  \left( \bm{J}^\top_{g}  \bm{\Sigma}^{-1}_{g} \bm{J}_{g} \right)^{-1}   
\end{equation}
where \\
 $\bm{J}_{g} = \frac{\partial g }{\partial \bm{x''} }  =  
\begin{bmatrix}
\begin{array}{ccc}
\frac{\partial h }{\partial \bm{x}_{c} } & \frac{\partial h }{\partial \bm{x}_{p} } & \bm{0}
\\  
\frac{\partial f }{\partial \bm{x}_{c} } & \bm{0} & \frac{\partial f }{\partial \bm{x}_{p'} } 
\end{array}
\end{bmatrix}
= 
\begin{bmatrix}
\begin{array}{ccc}
\bm{J}_{A} & \bm{J}_{B} & \bm{0}
\\   
\bm{J}_{C} & \bm{0} & \bm{J}_{D} 
\end{array}
\end{bmatrix}
 $. Next, we derive $
\bm{J}^\top_{g}  \bm{\Sigma}^{-1}_{g} \bm{J}_{g}  
=
\begin{bmatrix}
\begin{array}{c:cc}
\bm{J}^\top_{A}\bm{\Sigma}^{-1}_{h}\bm{J}_{A} + \bm{J}^\top_{C}\bm{\Sigma}^{-1}_{f}\bm{J}_{C} & \bm{J}^\top_{A}\bm{\Sigma}^{-1}_{h}\bm{J}_{B} & \bm{J}^\top_{C}\bm{\Sigma}^{-1}_{f}\bm{J}_{D}
\\   \hdashline
\bm{J}^\top_{B}\bm{\Sigma}^{-1}_{h}\bm{J}_{A} & \bm{J}^\top_{B}\bm{\Sigma}^{-1}_{h}\bm{J}_{B} & \bm{0}
\\
\bm{J}^\top_{D}\bm{\Sigma}^{-1}_{f}\bm{J}_{C} & \bm{0} & \bm{J}^\top_{D}\bm{\Sigma}^{-1}_{f}\bm{J}_{D}
\end{array}
\end{bmatrix} 
=
\begin{bmatrix}
\begin{array}{c:cc}
\bm{H}_{A} + \bm{H}'_{A} & \bm{H}_{B} & \bm{H}'_{B}
\\   \hdashline
\bm{H}^\top_{B} & \bm{H}_{C} & \bm{0}
\\
\bm{H}'^\top_{B} & \bm{0} & \bm{H}'_{C}
\end{array}
\end{bmatrix} 
  $.

Similarly, it can be deduced that, under Gaussian noise assumption, the MLE of $\bm{x}_{c}$ has covariance:
\begin{equation}
\label{C3}
\bm{C}_{g} = \left( \bm{H}_{A} + \bm{H}'_{A} - \bm{H}_{B} \bm{H}^{-1}_{C} \bm{H}_{B}^\top - \bm{H}'_{B} \bm{H}'^{-1}_{C} \bm{H}'^\top_{B}  \right) ^{-1}
\end{equation}

With (\ref{C1}), (\ref{C2}) and (\ref{C3}), the following relationship is derived:
\begin{equation}
\bm{C}_{g}^{-1} = \bm{C}_{h}^{-1} + \bm{C}_{f}^{-1} 
\end{equation}
If matrices $ \bm{M} $ and $ \bm{M}' $ are real symmetric and $ \bm{M}-\bm{M}' $ is positive definite, we can define $ \bm{M} \succ \bm{M}' $. Hence, the covariance matrices $\bm{C}_{h}$ and $\bm{C}_{f}$ satisfy: $\bm{C}_{h} \succ \bm{0}$, $\bm{C}_{f} \succ \bm{0}$.

According to matrix theory,  we have: 
\begin{equation}
\bm{C}^{-1}_{g} \succ \bm{C}^{-1}_{h}  \Rightarrow \bm{C}_{h} \succ \bm{C}_{g} , \bm{C}^{-1}_{g} \succ \bm{C}^{-1}_{f}  \Rightarrow \bm{C}_{f} \succ \bm{C}_{g}
\end{equation}
which means the $i$-th largest eigenvalue of $\bm{C}_{g}$ is smaller than the $i$-th largest eigenvalue of $\bm{C}_{h}$ and $\bm{C}_{f}$ \cite{horn2012matrix}.

\subsection{ Uncertainty Analysis and Line Guidance Features}

According to (25), the following conclusions can be drawn: first, adding more new landmarks with their observations in the local BA leads to smaller uncertainty in the MLE of keyframe poses; second, the more accurate observations of the new landmarks, the smaller uncertainty in the MLE of keyframe poses, that is, the smaller $\bm{C}_{f}$, the smaller $\bm{C}_{g}$.

In our system, we sample $N$ points on a 2D line feature and back-project them to generate the map line landmarks. Each map line is represented by $N$ 3D points (named line guidance features).
Let $\bm{\theta}$ be the states (including local keyframe poses and landmark positions) to be estimated in the local BA. The local BA aims to minimize the reprojection errors between the projections of the 3D landmarks and their corresponding 2D observations in the local keyframes:   
\begin{equation} 
\mathop{\arg\min}_{\bm{\theta}}  \mathop{\sum}_{i\in\mathcal{K}}\left[    \mathop{\sum}_{j\in\mathcal{P}}\bm{e}^\top_{ij}\bm{\Sigma}^{-1}_{\bm{e}_{ij}} \bm{e}_{ij}  
+ \mathop{\sum}_{k\in\mathcal{L}}\bm{e}^\top_{ik}\bm{\Sigma}^{-1}_{\bm{e}_{ik}} \bm{e}_{ik}  \right]
\end{equation}
where $\mathcal{K}$, $\mathcal{P}$ and $\mathcal{L}$ respectively refer to the sets of local keyframes, local map points, local map lines  and  	$ \bm{e}_{ik} =[\bm{e}^\top_{ik1}, \cdots, \bm{e}^\top_{ikN}]^\top.  
 $

The expression of reprojection error $\bm{e}_{ij}$ is similar to (7). 
The observation parameterization of an ORB feature $\bm{P}_{ij}$ is defined as 
$ \bm{P}_{ij} = [ u, v, u-\frac{f_{x}b}{d} ]^\top   $ based on original pixel measurement $\bm{p}_{uv}=[u,v]^\top$ and depth measurement $d$, where $b$ is the baseline of camera.
Assuming pixel coordinates and $d$ satisfy the zero-mean Gaussian distribution with the standard deviation $\sigma_{p}$ and $\sigma_{d}$, the covariance of original measurements is: 
\begin{equation}
	cov(\bm{p}_{uv},d) = 
	\begin{bmatrix}
	\sigma_{p}^2 & 0 & 0  \\
	0 & \sigma_{p}^2 & 0   \\
	0 & 0 & \sigma_{d}^2 
	\end{bmatrix} 
\end{equation}
where $\sigma_{p}$ is related to the extraction layer of ORB feature in the Gaussian pyramid, and $\sigma_{d}$ is modeled by a quadratic function of $d$ \cite{smisek20133d}.
The covariance of $\bm{P}_{ij}$ is defined as: 
\begin{equation}
\bm{\Sigma}_{\bm{e}_{ij}} = \bm{J}_{ij} cov(\bm{p}_{uv},d) \bm{J}_{ij}^\top
\end{equation}
where $ \bm{J}_{ij} = \frac{\partial \bm{P}_{ij} }{\partial (\bm{p}_{uv}, d) } 
= 
	\begin{bmatrix}
	1 & 0 & 0  \\
	0 & 1 & 0   \\
	1 & 0 & \frac{f_{x}b}{d^2} 
	\end{bmatrix} 
$.

The reprojection error $\bm{e}_{iks} (s=1,\cdots,N) $ is obtained by the distance between the projected 2D position of 3D line-guidance feature $\bm{P}_{iks} $ and the corresponding straight line $\bm{l}_{ik}$ (i.e. observation) in the image plane:
\begin{equation}
\bm{e}_{iks} = \bm{l}^\top_{ik} \cdot \pi \left(\bm{T}_{i}(\bm{P}_{iks}) \right) 
\end{equation}

Compared to point-based methods, our method is based on point and line guidance features, and expands the number of landmarks to be estimated in the local BA, which will result in smaller estimation uncertainty of keyframe poses according to the uncertainty analysis in this section.  Line features are often at the edges of objects, such that the corresponding  depth measurements are noisy. Considering the noisy depth measurements and the mismatch of line features, oversampling points on a line will introduce too many landmarks with poor initial values and have a bad impact on the optimization problem. A good SLAM algorithm should carefully balance the quantity and quality of landmarks in the bundle adjustment.


\section{Experimental Validation}

In this section, we compare our algorithm with several state-of-the-art visual SLAM methods on two public RGB-D datasets: 

\begin{itemize}
	\item ICL-NUIM \cite{handa2014benchmark} is a dataset providing RGB and depth image sequences in two synthetically generated scenes (the living room and office room).  Some sequences is challenging to estimate camera poses due to low-textured characteristics of the environments. 
	
	\item TUM RGB-D benchmark \cite{sturm2012benchmark} contains sequences from  RGB-D sensors with different texture, structure and illumination conditions in real indoor environments. 
\end{itemize}

We implement our line guidance feature-based method by sampling five points for each line. The difference in the number of sampling points can affect the performance of our algorithm, which will be discussed later. All experiments were carried out with an Intel Core i7-8550U (4 cores @4.0 GHz) and 8Gb RAM.

\subsection{ICL-NUIM Dataset}

We first compare the proposed method against some state-of-the-art approaches, including ORB-SLAM2, DVO-SLAM \cite{kerl2013dense} and L-SLAM \cite{kim2018linear}. Artificial noise was used in the image data to simulate realistic sensor noise in this dataset, meanwhile every sequence has a version without noise. 

The evaluation results are shown in TABLE \uppercase\expandafter{\romannumeral1} for all the sequences with noise. The  Absolute Trajectory Error (ATE) is used for measuring the root mean squared error (RMSE) between the estimated trajectory and ground truth. Though L-SLAM performs best in three noise sequences, it is limited to planar environments. Note that our method outperforms the other methods in half of all the noise sequences. We show the 788th frame (low-textured scene) from sequence \textit{lr-1} during the tracking process in Fig.4(a) and the estimated trajectories comparison between ORB-SLAM2 and Ours in Fig. 7. In addition, we also conduct experiments for ORB-SLAM2 and Ours in all the sequences without noise, as shown in TABLE \uppercase\expandafter{\romannumeral1}. According to the results of ORB-SLAM2 and Ours in TABLE \uppercase\expandafter{\romannumeral1}, it can be found that the higher quality of newly added landmarks with their measurements leads to more accurate pose estimation, which is basically consistent with our analysis in section V-D.

\begin{figure}[htbp]
	\centering
	\subfigure[]{
		\begin{minipage}[t]{0.38\linewidth}
		    \centering
			\includegraphics[width=1.36in]{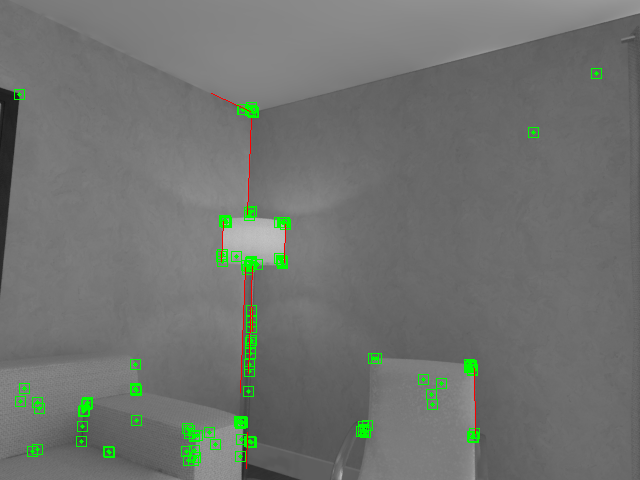}
		\end{minipage}
	}
	\subfigure[]{
		\begin{minipage}[t]{0.4\linewidth}
			\centering
			\includegraphics[width=1.66in]{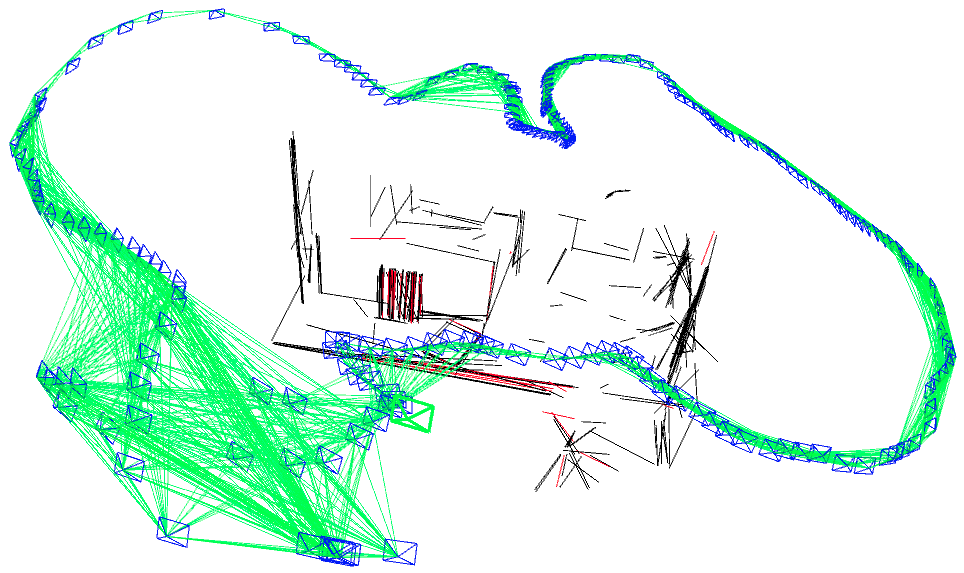}
		\end{minipage}
	}
	\centering
	\caption{ $ \mathbf{(a)} $ The 788th frame (low-textured scene) from sequence \textit{lr-1} in the tracking process. Green squares denote the tracked point features, and red indicate the tracked line features. $ \mathbf{(b)} $  A map with line features in \textit{fr3/office}. }
\end{figure}

\begin{table}[htbp]
	\caption{ Results of ATE (unit: $m$) in the ICL-NUIM Benchmark }
	\begin{center}
		\begin{tabular}{|c|c|c|c|c|c|c|}
			\hline
			& \multicolumn{2}{|c|}{ Without noise } & \multicolumn{4}{|c|}{ With noise } \\
			\hline
			Seq. & Ours & \makecell[t]{ORB- \\ SLAM2} & Ours & \makecell[t]{ORB- \\ SLAM2} &  \makecell[t]{L- \\ SLAM} & \makecell[t]{DVO- \\ SLAM}  \\
			\hline
			lr-0 & \textbf{0.005} & 0.008 & 0.009 & \textbf{0.008} & 0.012 & 0.108  \\
			\hline
			lr-1 & \textbf{0.008} & 0.126 & \textbf{0.009} & 0.134 & 0.027 & 0.059   \\
			\hline
			lr-2 & \textbf{0.016} & 0.023 & \textbf{0.016} & 0.032 & 0.053 & 0.375   \\
			\hline
			lr-3 & \textbf{0.007} & 0.017 & \textbf{0.012} & 0.014 & 0.143 & 0.433    \\
			\hline
			of-0 & \textbf{0.019} & 0.030 & 0.040 & 0.054 & \textbf{0.020} & 0.244    \\
			\hline    
			of-1 & \textbf{0.017} & 0.051 & 0.025 & 0.058 & \textbf{0.015} & 0.178    \\
			\hline
			of-2 & \textbf{0.014} & 0.015 & \textbf{0.021} & 0.025 & 0.026 & 0.099    \\
			\hline    
			of-3 & \textbf{0.008} & 0.070 & 0.015  & 0.050 & \textbf{0.011} & 0.079  \\
			\hline
			
		\end{tabular}
	\end{center}
\end{table}

\subsection{TUM RGB-D Dataset}

We choose \textit{fr3/office}  for mapping visualization, as shown in Fig. 4(b). Line features can provide more spatial structural information of the environments than point features. Fig. 5 shows the difference of sparse map between ORB-SLAM2 and Ours, which indicates that our method can construct a more accurate map structure. 

In this dataset, we compare our algorithm with eight state-of-the-art SLAM methods: ORB-SLAM2, an indirect point-line-based method PL-SLAM \cite{pumarola2017pl}, a direct point-line-based method DLGO \cite{li2018direct}, RKD-SLAM \cite{liu2017robust}, Kintinuous \cite{whelan2012kintinuous}, ElasticFusion \cite{whelan2016elasticfusion}, DVO-SLAM and RGBD SLAM \cite{endres20133}.


\begin{figure}[tbp]
	\centering
	\subfigure[fr3/nstr\_tex\_far]{
		\begin{minipage}[t]{0.5\linewidth}
			\centering
			\includegraphics[width=1.5in]{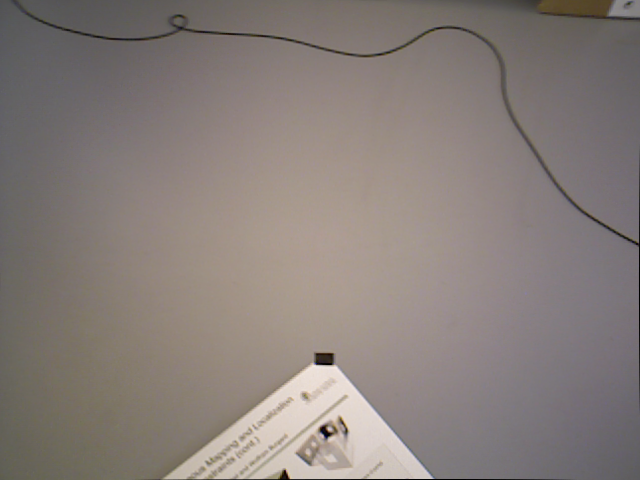}
		\end{minipage}
	}
	\subfigure[ORB-SLAM2 with points]{
		\begin{minipage}[t]{0.4\linewidth}
			\centering
			\includegraphics[width=1.5in]{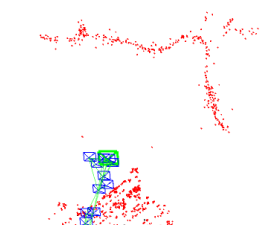}
		\end{minipage}
	}
	\subfigure[Ours with points]{
		\begin{minipage}[t]{0.5\linewidth}
			\centering
			\includegraphics[width=1.5in]{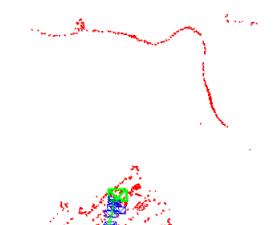}
		\end{minipage}
	}
	\subfigure[Ours with lines]{
		\begin{minipage}[t]{0.4\linewidth}
			\centering
			\includegraphics[width=1.5in]{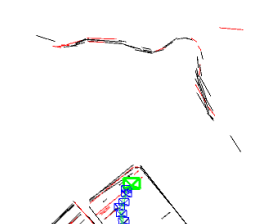}
		\end{minipage}
	}
	\centering
	\caption{ Difference of sparse map between ORB-SLAM2 and Ours.  }
\end{figure}

The ATE results are shown in  TABLE \uppercase\expandafter{\romannumeral2}. From the table, we can find that: (i) the methods with combination of points and lines lead to higher accuracy than those only using points; (ii) our method achieves better performance than other methods in more than half of the sequences. In dynamic environments (\textit{fr3/sit\_static}, \textit{fr3/sit\_half}, \textit{fr3/walk\_half}), the performance of ORB-SLAM2 is worse than the point-line-based methods, since the line features are basically on the static objects. PL-SLAM fails in the sequence \textit{fr3/nstr\_tex\_far} and the main reason is that the images are blurred due to the fast camera motion.  
Additionally, we show the estimated trajectories (ORB-SLAM2 and Ours) corresponding to the sequences \textit{fr1/room} and \textit{fr3/sit\_half}  in Fig. 7. The performance of ORB-SLAM2 drops in the low-textured (\textit{lr-1}) and low dynamic (\textit{fr3/sit\_half}) environments. We also show the comparison of trajectory errors in different axes for sequence \textit{fr1/360} in Fig. 6.

\begin{figure}[thpb]
	\centering
	\includegraphics[width=0.415\textwidth]{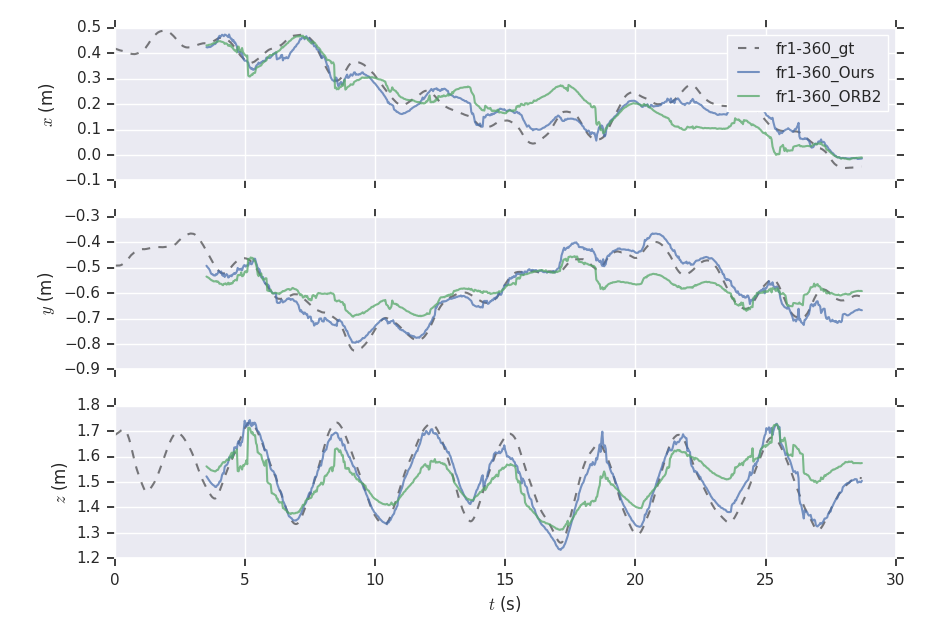}
	\caption{The xyz-axis trajectory errors of ORB-SLAM2 and Ours for sequence \textit{fr1/360}. "gt" indicates ground truth, and "ORB2" indicates ORB-SLAM2.    }
\end{figure}

\begin{figure*}[htbp]
	\centering
	\subfigure[lr-1 with noise]{
		\begin{minipage}[t]{0.23\linewidth}
			\centering
			\includegraphics[width=1.8in]{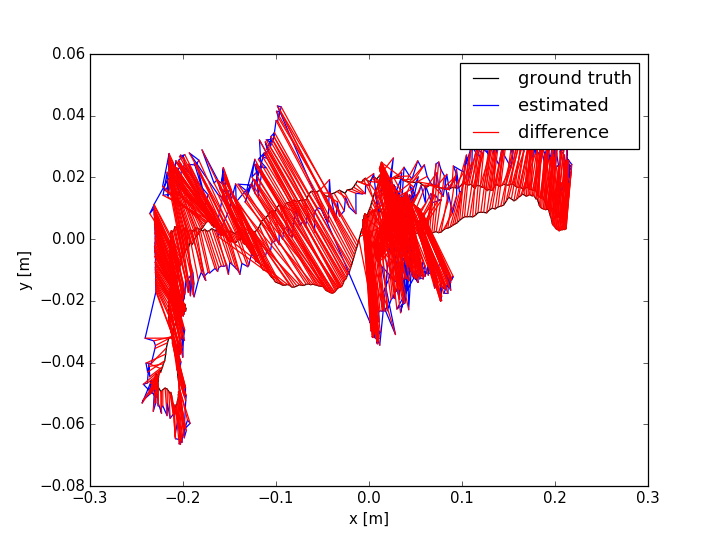}
		\end{minipage}%
	}%
	\subfigure[of-3 with noise]{
		\begin{minipage}[t]{0.23\linewidth}
			\centering
			\includegraphics[width=1.8in]{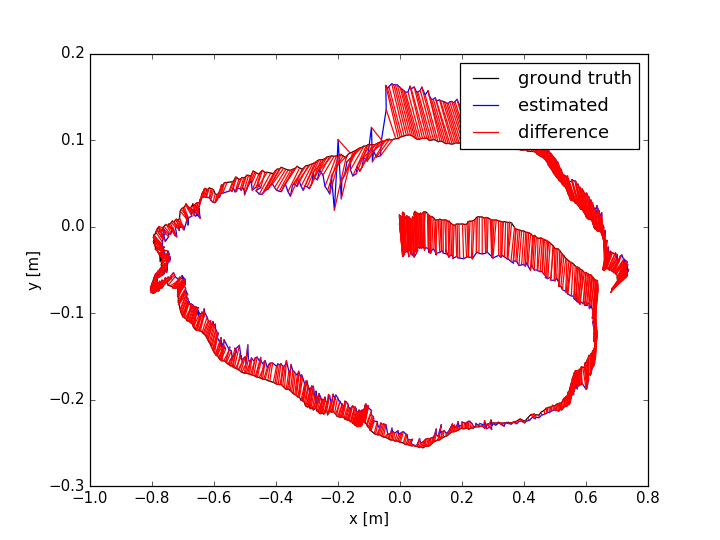}
		\end{minipage}%
	}  
	\subfigure[fr1/room]{
		\begin{minipage}[t]{0.23\linewidth}
			\centering
			\includegraphics[width=1.8in]{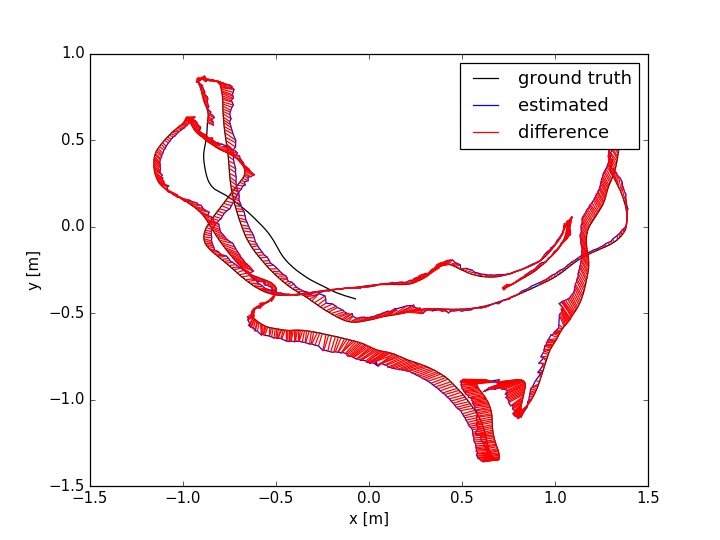}
		\end{minipage}%
	}  
	\subfigure[f3/sit\_half]{
		\begin{minipage}[t]{0.23\linewidth}
			\centering
			\includegraphics[width=1.8in]{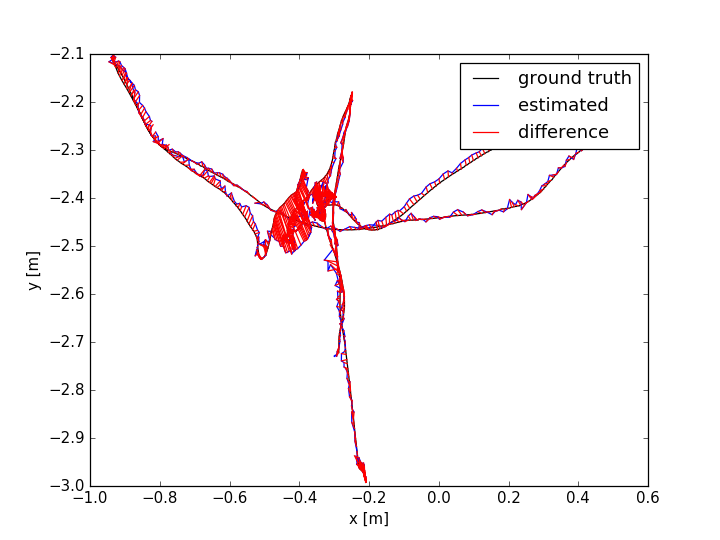}
		\end{minipage}%
	}  
	\quad
	\subfigure[lr-1 with noise]{
		\begin{minipage}[t]{0.23\linewidth}
			\centering
			\includegraphics[width=1.8in]{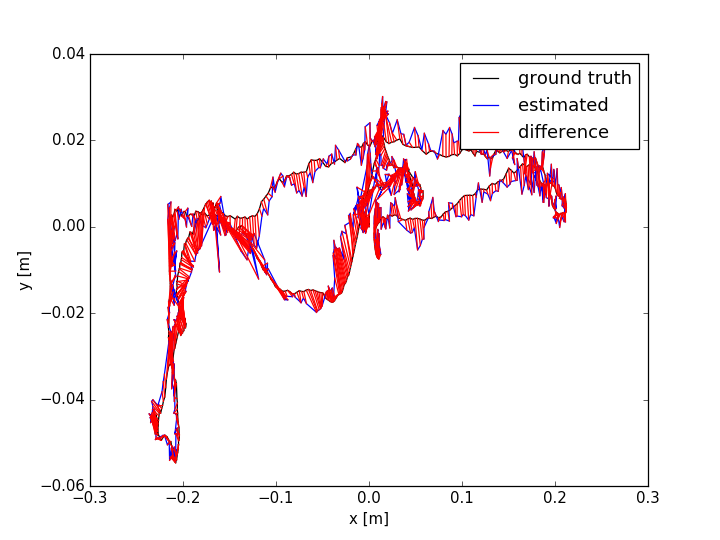}
		\end{minipage}
	}%
	\subfigure[of-3 with noise]{
		\begin{minipage}[t]{0.23\linewidth}
			\centering
			\includegraphics[width=1.8in]{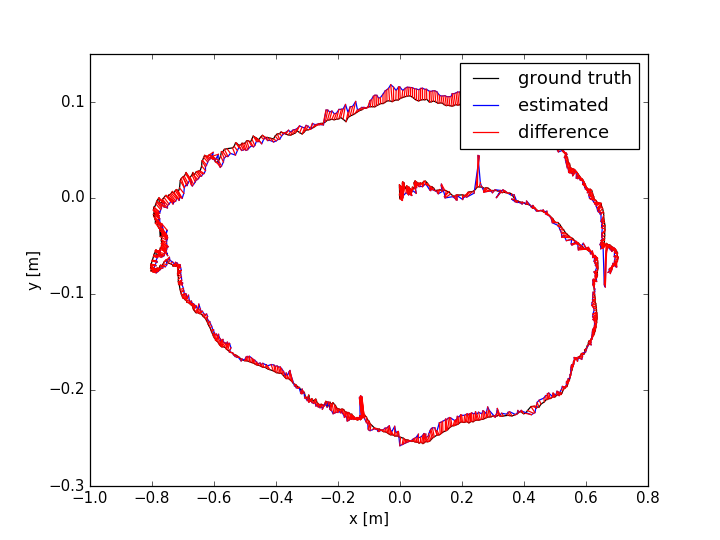}
		\end{minipage}
	}%
	\subfigure[fr1/room]{
		\begin{minipage}[t]{0.23\linewidth}
			\centering
			\includegraphics[width=1.8in]{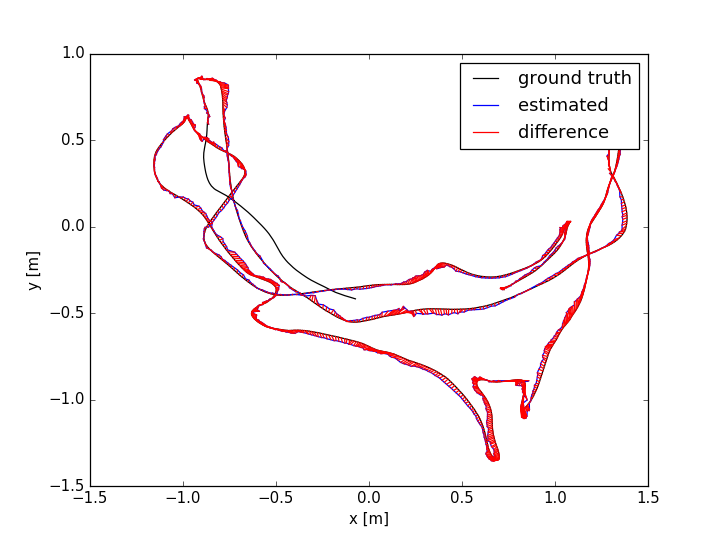}
		\end{minipage}%
	}  
	\subfigure[fr3/sit\_half]{
		\begin{minipage}[t]{0.23\linewidth}
			\centering
			\includegraphics[width=1.8in]{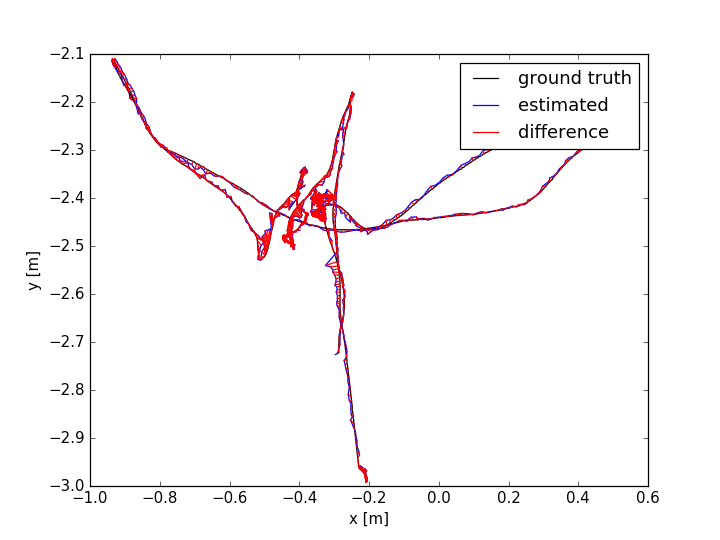}
		\end{minipage}%
	}  
	\centering
	\caption{ Comparison of estimated trajectories between ORB-SLAM2 and Ours. The first row is the results of ORB-SLAM2, while the second is Ours. Due to low-texture (\textit{lr-1}) or low dynamic objects (\textit{fr3/sit\_half}), the estimated poses vary dramatically in some locations for ORB-SLAM2. All the trajectories are drawn by the official evaluation tool on the TUM RGB-D dataset.     }
\end{figure*}

\begin{table*}[ht]
	\caption{Comparison of the ATE results (unit: $m$) among different methods in the TUM RGB-D dataset}
	\label{table_example}
	\begin{center}
		\begin{tabular}{|c|c|c|c|c|c|c|c|c|c|}
			\hline
			Sequence & Ours & ORB-SLAM2 & PL-SLAM & DLGO & RKD-SLAM & Kintinuous & ElasticFusion & DVO-SLAM & RGBD SLAM   \\
			\hline
			fr1/xyz & 0.009 & 0.010 & 0.012 & 0.054 & \textbf{0.007} & 0.017 & 0.011 & 0.011 & -  \\
			\hline
			fr1/desk2 & \textbf{0.022} & 0.024 & - & - & 0.024 & 0.071 & 0.048 & 0.046 & -  \\
			\hline
			fr1/floor & \textbf{0.013} & 0.016 & 0.076 & - & 0.262 & - & - & - & -   \\
			\hline
			fr1/room & \textbf{0.030} & 0.059 & - & - & 0.134 & 0.075 & 0.068 & 0.053 & 0.087   \\
			\hline    
			fr1/360 & \textbf{0.068} & 0.228 & - & - & 0.109 & - & 0.108 & 0.083 & -   \\
			\hline
		 
			fr2/desk & \textbf{0.008} & 0.009 & - & 1.33 & 0.071 & 0.034 & 0.071 & 0.017 & 0.057  \\
			\hline
			fr2/desk\_person & 0.007 & \textbf{0.006} & 0.020 & 0.412 & 0.045 & - & - & - & -   \\
			\hline
			fr3/office & 0.010 & \textbf{0.010} & 0.020 & 1.168 & 0.028 & 0.030 & 0.017 & 0.035 & -   \\
			\hline 
			fr3/nstr\_tex\_far & \textbf{0.023} & 0.051 & X & 0.575 & 0.053 & - & 0.074 & - & -   \\
			\hline
			fr3/nstr\_tex\_near & \textbf{0.013} & 0.024 & 0.021 & 0.060 & 0.027 & 0.031 & 0.016 & 0.018 & -   \\
			\hline
			fr3/str\_tex\_far & 0.010 & 0.011 & \textbf{0.009} & 0.744 & 0.016 & - & 0.013 & - & -   \\
			\hline
			fr3/str\_tex\_near & \textbf{0.009} & 0.011 & 0.013 & - & 0.018 & - & 0.015 & - & -   \\
			\hline
			fr3/sit\_static & \textbf{0.006} & 0.009 & - & - & 0.009 & - & - & - & -  \\
			\hline
			fr3/sit\_half & \textbf{0.011} & 0.021 & 0.013 & 0.160 & 0.019 & - & - & - & -    \\
			\hline
			fr3/walk\_half & 0.091 & 0.431 & \textbf{0.016} & 0.374 & 0.182 & - & - & - & -  \\
			\hline
		\end{tabular}
	\end{center}
\end{table*}

\subsection{Number of Sampling Point}

We test the ATE of our method with different number of sampling points and the same parameter setting in the sequences  \textit{fr1/room}, \textit{fr3/nstr\_tex\_far}, \textit{lr-1\_noise} and \textit{of-3\_noise}, and the results are shown in Fig. 8. 
In our opinion, the quantity and quality of newly added landmarks need to be balanced and paid enough attention.  Sampling too many points will lead  to  reduction of the quality of line landmarks due to the noisy depth measurements and result in an increase of the number of outliers. It would be better to sample a small number of points and we choose five sampling points in our implementation. 

\begin{figure}[thpb]
	\centering
	\includegraphics[width=0.39\textwidth]{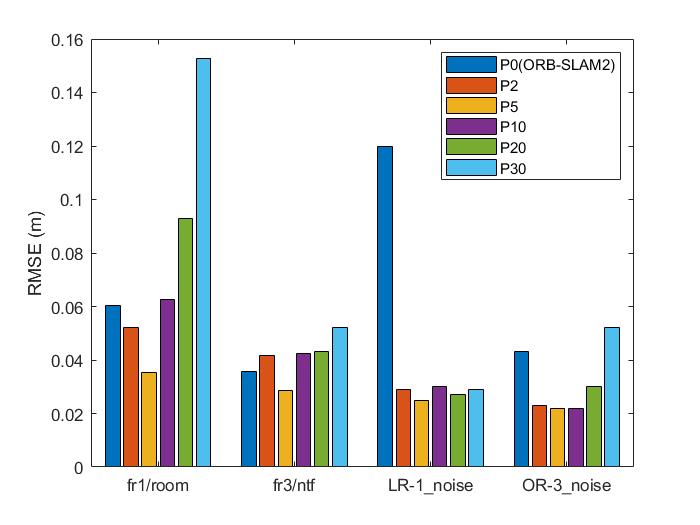}
	\caption{ The results under different number of sampling points. P0 refers to ORB-SLAM2, and P$X$ means sampling $X$ points on a line. }
\end{figure}

\subsection{Computation Time}

The point-line-based visual SLAM methods improve the accuracy and robustness, but at the same time lead to higher computational complexity. In particular, the extraction and matching of line features increase a lot of time consumption. We summarize the results of mean tracking time per frame for ORB-SLAM2 and Ours in  TABLE \uppercase\expandafter{\romannumeral3}.  

\begin{table}[ht]
	\centering
	\caption{ Comparison of mean tracking time (unit: $ms$) }
	\begin{tabular}{ccccccc} 
		\toprule
		\toprule
		Method  & fr1/desk2 & fr1/360  & fr2/desk  & fr3/office & fr3/ntn   \\
		\midrule
		ORB-SLAM2  & 53.0 & 44.2 &  61.5  & 58.7 & 47.3    \\
		
		Ours  &  71.7 & 63.9 & 78.5  & 83.0 & 75.3  \\
		
		\bottomrule
		\bottomrule
	\end{tabular}
\end{table}

\section{CONCLUSIONS}

In this work, we proposed a RGB-D SLAM method through combining both point and line features, which allows to improve the accuracy and robustness in some situations where point-only based methods are prone to fail due to not enough feature points. We have proved that, compared with original landmarks, adding new landmarks with corresponding independent observations can ensure smaller uncertainty in the estimation of keyframe poses in the local bundle adjustment. Compared to several state-of-the-art methods, the proposed method improves the accuracy and robustness under challenging environments, such as the scenes with low-texture, fast camera movement and low dynamic objects.  In the future, we will further optimize the system to increase the time efficiency of point-line-based SLAM method.










\bibliographystyle{IEEEtran}  
\bibliography{IEEEabrv,reference}

\end{document}